\documentclass[tog]{acmsiggraph}

\def\BibTeX{{\rm B\kern-.05em{\sc i\kern-.025em b}\kern-.08em
    T\kern-.1667em\lower.7ex\hbox{E}\kern-.125emX}}

\TOGonlineid{45678}

\TOGvolume{0}
\TOGnumber{0}

\title{GarmentGAN: Photo-realistic Adversarial Fashion Transfer}

\author{Amir Hossein Raffiee\thanks{e-mail:amir.raffiee@salesforce.com}, Michael Sollami\thanks{e-mail:msollami@salesforce.com}\\Salesforce Einstein}
\pdfauthor{author1}


\usepackage{amssymb}
\usepackage{amsmath}

\usepackage[utf8]{inputenc}
\usepackage[table]{xcolor}

\begin{document}


 \teaser{
   \includegraphics[height=3.3in]{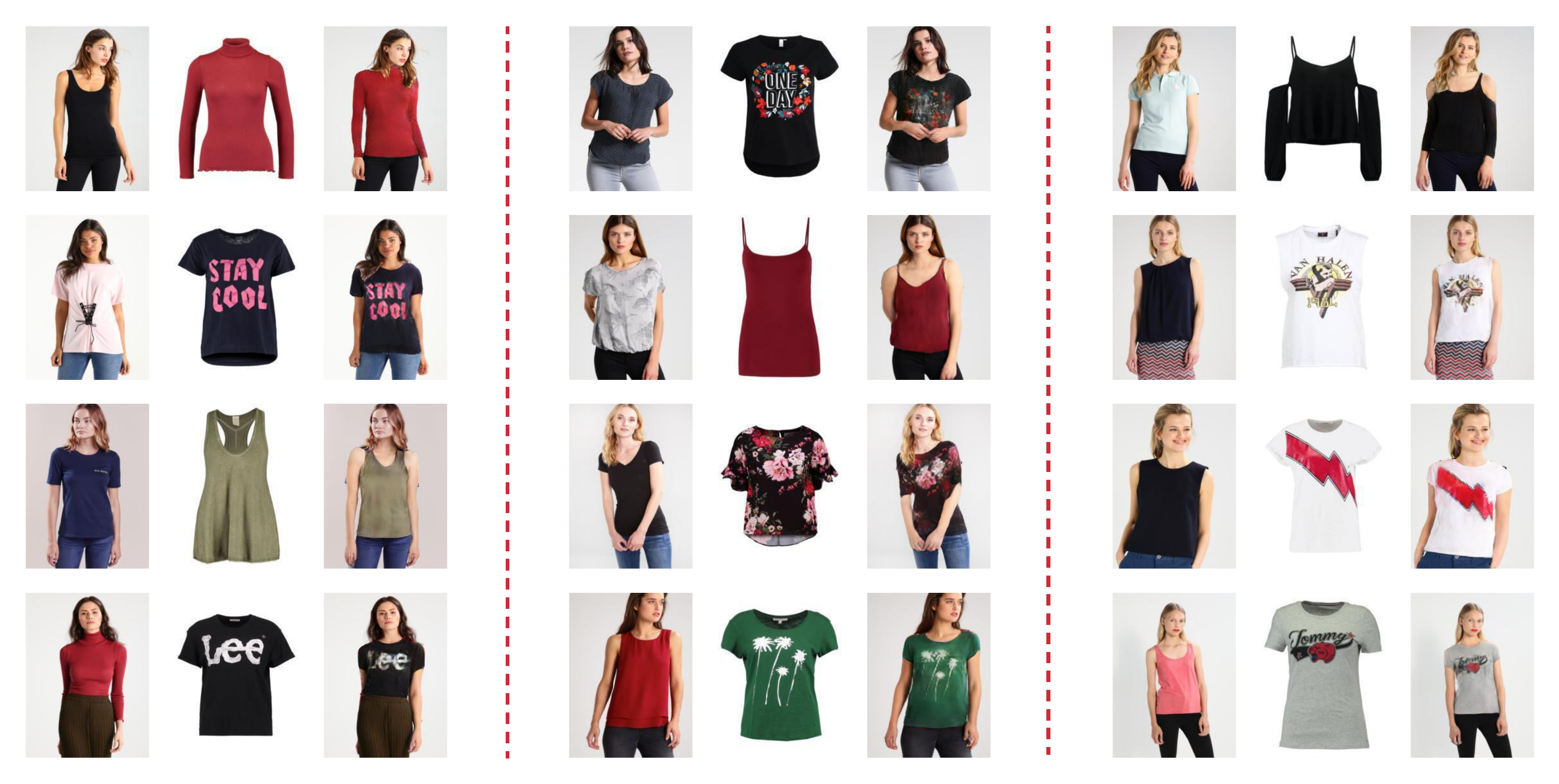}
   \caption{The model proposed in this work transfers the target clothing item (middle columns) onto a reference body (leftmost columns) generating photo-realistic images (rightmost columns) at higher resolution and fidelity than other state-of-the-art methods. The results in the above grid illustrate the preservation of shapes and textures during the image generation process over multiple clothing types and arbitrary poses.}
   \label{front-image}
 }

\maketitle

\begin{abstract}

The garment transfer problem comprises two tasks: learning to separate a person's body (pose, shape, color) from their clothing (garment type, shape, style) and then generating new images of the wearer dressed in arbitrary garments. We present GarmentGAN, a new algorithm that performs image-based garment transfer through generative adversarial methods. The GarmentGAN framework allows users to virtually try-on items before purchase and generalizes to various apparel types. GarmentGAN requires as input only two images, namely, a picture of the target fashion item and an image containing the customer. The output is a synthetic image wherein the customer is wearing the target apparel. In order to make the generated image look photo-realistic, we employ the use of novel generative adversarial techniques \cite{kubo2018generative}. GarmentGAN improves on existing methods in the realism of generated imagery and solves various problems related to self-occlusions. Our proposed model incorporates additional information during training, utilizing both segmentation maps and body key-point information. We show qualitative and quantitative comparisons to several other networks to demonstrate the effectiveness of this technique.

\end{abstract}





\section{Introduction}

Although online apparel shopping has grown significantly in the last decade, many consumers still prefer brick and mortar experiences where they can inspect and try-on products before purchase. This trend has attracted the attention of researchers seeking to make online shopping environments more convenient. To this end, we develop a virtual dressing room for e-commerce, thus enabling customers to try out an arbitrary number of fashion items without physically wearing them.

Among the most recent image synthesis approaches developed for virtual try-on purposes, \cite{han2018viton,wang2018toward} employed course-to-fine networks to generate the image of a person conditioned on a cloth of interest. However, since these methods do not use full body parse information, the resulting images are mostly blurry and in some cases unrealistic. To address the indicated deficiency \cite{dong2019towards,issenhuth2019end} leveraged the body part information, but their method fails to produce realistic images when the human poses are even slightly complicated (e.g. when hands or arms  occlude other parts of the body) which limits the application of the proposed algorithms for photo-realistic images.

Furthermore, none of the recently published models that we surveyed are able to faithfully transfer finer scale details of clothing such as small text or logos. Other existing approaches \cite{lahner2018deepwrinkles,pons2017clothcap,zhang2017detailed} use 3D models to resolve such deficiencies associated with existing methods. However, massive computation and significant labor costs required for collecting 3D annotated data restrict their applications in real-time virtual try-on systems.

In this paper, we propose an image-based virtual try-on algorithm conditioned on the images of a given garment and a person to fit the cloth. This technique aims to learn a mapping between the input images of a reference person and an arbitrary clothing item to the image of the same person wearing the target clothing. This problem has been studied widely in previous works \cite{han2018viton,wang2018toward,dong2019towards,issenhuth2019end,raj2018swapnet,liu2019swapgan}, however, all these proposed methods still suffer from the limitations mentioned above. Our proposed method, GarmentGAN, is a multi-stage framework that generates a photo-realistic image of a consumer wearing any desired outfit, wherein the chosen garments are fit seamlessly onto the person's body and details are maintained down to the single pixel level. 

In the first stage of this framework, a shape transfer network synthesizes a segmentation map associated with a person wearing the target clothing item. The output of this network provides the layout information and body part boundaries that guide the synthesis of the final image. In the second stage, a Thin-Plate-Spline transformation warps the desired cloth to address the misalignment between arbitrary body pose and the input clothing item. This warped cloth image is then passed through an appearance transfer network that generates the final RGB-colorspace image of the reference person wearing the desired cloth preserving fine details in the identity of the clothing item (e.g. logos or patterns).

To show the performance of our model, we conduct multiple experiments on the fashion dataset collected by \cite{han2018viton} and provide quantitative and qualitative comparisons with state-of-the-art methods. This approach synthesizes photographs at high levels of realism by preserving details for both wearer and target garment as illustrated in Fig. \ref{front-image}. 

We now summarize the contributions of GarmentGAN:
\begin{itemize}
  \item The system comprises two separate GANs: a shape transfer network and an appearance transfer network
  \item Uses a geometric alignment module that is trained in an End-to-End fashion 
  \item Handles complex body pose, hand gestures, and occlusions with a new method of masking semantic segmentation maps
  \item Resolves issues associated with the loss of target clothing characteristics (e.g. collar shape) during the garment transfer 
  \item Preserves the identity of the reference person and clothing items that should remain unchanged during the try-on task
\end{itemize}

In this work, we conduct various experiments to evaluate the performance of the model and provide a comprehensive discussion on observed results and comparisons with state-of-the-art models.

\section{Related Works}

\subsection{Image Synthesis}
Classical texture synthesis methods gave way in 2015 to generative adversarial networks (GANs) \cite{goodfellow2014generative}, which have impressive capabilities for synthesizing realistic content and are now the main component of any model aiming to generate images. These typically consist of two neural networks called the generator and discriminator, where the generator learns the distribution of real images by generating images that are indistinguishable from real images, and the discriminator learns to classify the images into real and fake. Existing methods employ GANs for various synthesis purposes such as text-to-image \cite{reed2016generative,zhu2017your} and image-to-image \cite{isola2017image,kim2017learning,yi2017dualgan} tasks.

\subsection{Person image generation}
Person image generation is another challenging task that has attracted the attention of researchers in recent years due to broad applicability. Skeleton-guided image generation \cite{yan2017skeleton} was among the first attempts to synthesize human motion conditioned on a single human image and a sequence of human skeleton information. PG2 \cite{ma2017pose} utilized a coarse-to-fine strategy to generate an image of a person at arbitrary poses given a single image of a person and the desired pose. This work was further enhanced by disentangling the image background from various body parts (foreground) \cite{ma2018disentangled} which guides the network to synthesize images at a new pose with higher quality. DeformableGANs \cite{siarohin2018deformable} also attempted to generate an image of a person in a novel pose using a deformable skip connection to handle the misalignment between the input and target pose. \cite{pumarola2018unsupervised} proposed a fully unsupervised generative model that uses the CycleGan \cite{zhu2017unpaired} approach for multi-view synthesize task. \cite{lassner2017generative} is among the first works that proposed full-body image generation by synthesizing human parsing maps and mapped those back into image space. Furthermore, \cite{chen2019unpaired} generates person images using the coarse sketch of their pose; this method provides either control over the outfit worn by the person or conditional sampling to the user.

\subsection{Virtual try-on}

n 2018, VITON \cite{han2018viton} proposed a method that transfers target clothing onto a person in a coarse-to-fine manner using shape context matching module to warp the target clothing item to match the pose of the reference person. Additionally, CP-VTON \cite{wang2018toward} employs a warping transformation module that  paints the target garment onto the reference body, however they leverage a composition network to integrate the warped clothes onto the person’s image. This practice shows significant improvement in preserving garment identity. Besides the previous works, CAGAN \cite{jetchev2017conditional} proposed a CycleGan-based \cite{zhu2017unpaired} framework that transfers an arbitrary garment onto a reference body without requiring any information regarding body pose or shape. Despite the recent progress cited in this section, all aforementioned synthesis methods suffer from qualitative defects such as blurriness and artifacts. Furthermore, these methods merely fit target clothes onto a fixed predetermined human pose. To address this shortcoming in particular, SwapNet \cite{raj2018swapnet} introduces a framework that transfers the clothes between two bodies. This method divides the image generation task into two sub-tasks: segmentation map synthesis and transference of the clothing characteristics onto the previously generated map. 
However the SwapNet framework has difficulty handling pose changes between source and target images and  cannot hallucinate details for occluded regions. It is at this point where we introduce GarmentGAN's contribution which overcomes these issues.

\begin{figure*}[ht]
    \centering
       \includegraphics[height=3.3in]{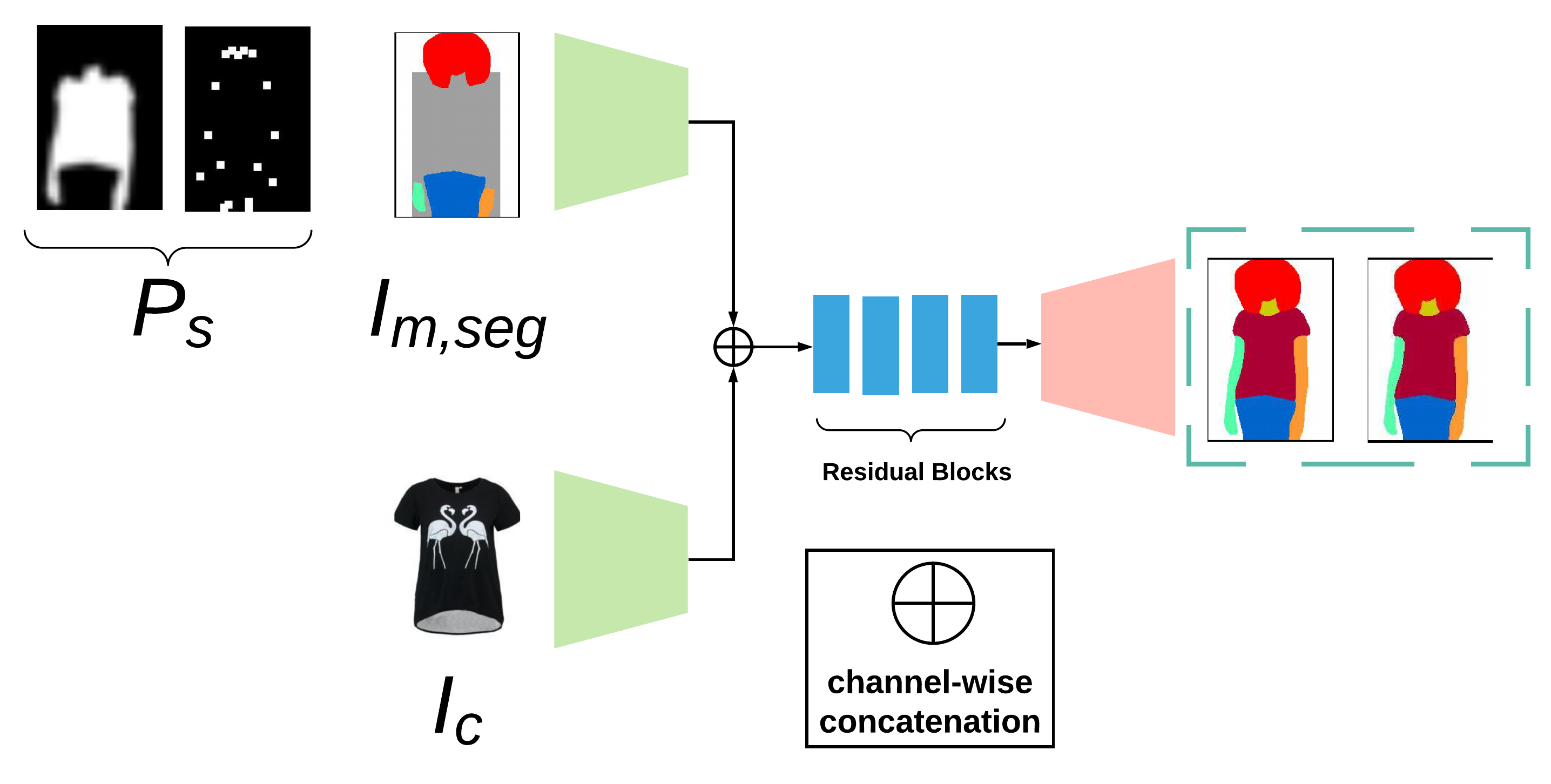}
\caption{Shape transfer network architecture }
    \label{seg_1}
\end{figure*}

\section{GarmentGAN's Approach}

The works referenced in the previous section lends support to the claim that garment transfer is a complex task which cannot be accomplished in any single-stage framework. Inspired by coarse-to-fine strategies adopted in previous works \cite{han2019compatible,wang2018toward} we propose a two-stage framework consisting of (i) shape transfer network that learns to generate a reasonable semantic map given the image of the person and desired fashion item and (ii) appearance transfer network that synthesizes a realistic RGB-colorspace image of the person wearing the garment while preserving finer semantic details.

\subsection{Shape transfer Network}
\begin{figure}[h]
    \centering
        \includegraphics[height=1.7in]{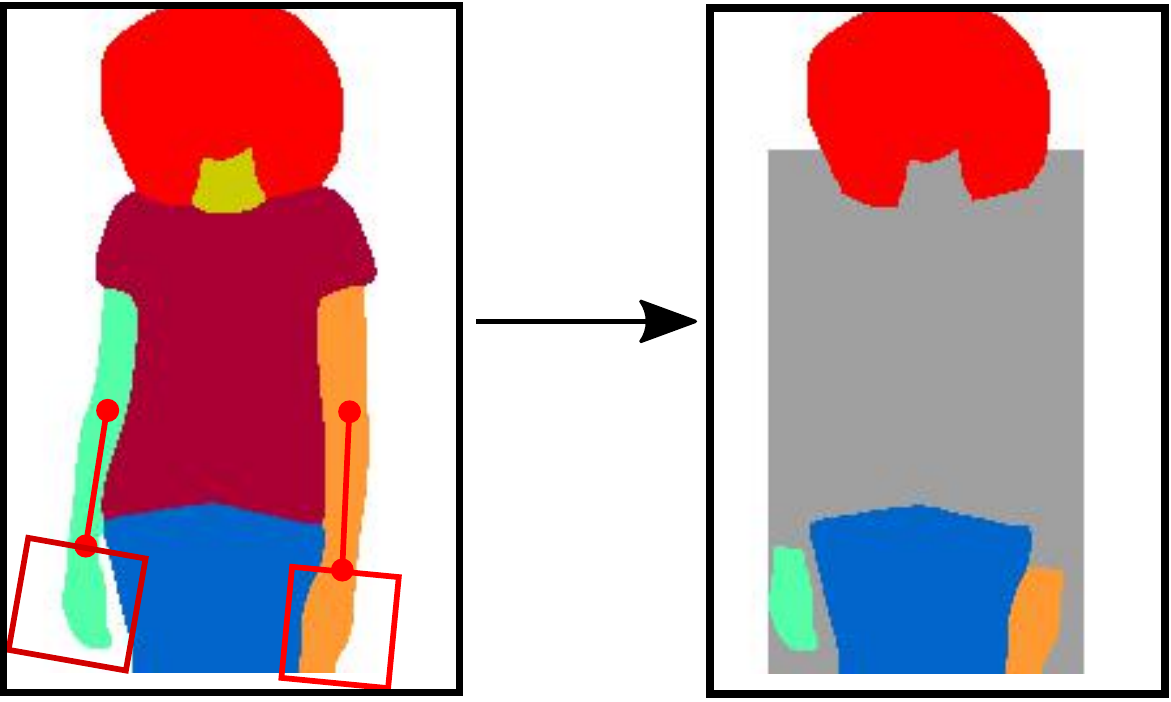}
\caption{Masking input segmentation map such that hand regions are retained}
    \label{seg_2}  
\end{figure}

In this section we explain the design of the proposed network for shape transference. The overview of the network is illustrated in Fig. \ref{seg_1}. As depicted in this figure, we require that the semantic segmentation map of the reference person be passed onto the network. Hence, a  semantic parser \cite{gong2018instance} is trained on the LIP dataset \cite{gong2017look} and utilized to extract the pixel-wise semantic map of the reference person ($I_{person}\in R^{H\times W\times 3}$). This unet model parses $I_{person}$ into ten different categories, namely $\{$\textit{hat}, \textit{face/hair}, \textit{torso}, \textit{left arm}, \textit{right arm}, \textit{top clothes}, \textit{bottom cloths}, \textit{legs}, \textit{shoes}, \textit{background}$\}$. The resulting map is further transformed into a 10-channel binary map ($I_{seg}\in \{0,1\}^{H\times W\times 10}$) that serves as ground truth during the training process. The proposed generator ($G_{shape}$) has an encoder-decoder architecture that synthesizes a new shape map ($\hat{I}_{seg}$) conditioned on person representation ($P_s$), masked segmentation map ($I_{m,seg}$) and the image of desired clothing ($I_{c}$).
\begin{equation}\label{eq:0}
\hat{I}_{seg} = G_{shape}(I_{m,seg},P_s,I_c)
\end{equation}
Note that in this version of GarmentGAN, we focus merely on transferring upper torso clothes, however, the proposed algorithm can be trivially generalized for other pieces of clothing (e.g. shoes, pants, etc.). In our method, the masked segmentation map $I_{m,seg}$ is obtained by masking the smallest rectangle that encompasses the target area, which in this case is torso, arms, and top clothes regions as shown in Fig. \ref{seg_2}. In order to retain the identity of the reference person during transference, the masking operation is only applied on the segmentation channels associated with a torso, arms, top clothes and background, and remaining channels are left unchanged. Furthermore, in order to make the network robust to arbitrary poses (the cases where the reference person has complicated gestures or cases in which self-occlusion occurs), we retain the semantic information of left and right hands in the input segmentation map depicted in Fig. \ref{seg_2}. In this regard, the pixels associated with hand regions are identified using the keypoint locations that characterize the arms, elbows, and wrists. Using these locations, we can find a square that has its sides normal to the connecting line between elbow and wrist keypoints. Subsequently, the pixels belonging to the hands part and falling in the obtained squares (shown by red color sides) are retained during masking operation and fed to the shape transfer network.
The information about the pose and the body shape of reference person is fed into the network through a  clothing-agnostic feature representation proposed in \cite{wang2018toward,han2018viton}. In this regard, any off-the-shelf pose estimator \cite{chen2018cascaded} (trained on MS-COCO) \cite{lin2014microsoft} may be employed to identify 17 keypoints of the body. The results are represented as a 17-channel binary map in which each channel corresponds to the heat map for each keypoint. Furthermore, the body shape is represented as a blurred binary mask of the arms, torso, and top clothes generated from the extracted segmentation map (as shown in Fig. \ref{seg_1}). Concatenating the aforementioned feature representations creates the person representation ($P_s \in {R^{H\times W\times 18}}$) which is then fed into the shape transfer network.
\begin{figure*}[ht]
    \centering
       \includegraphics[height=4.9in]{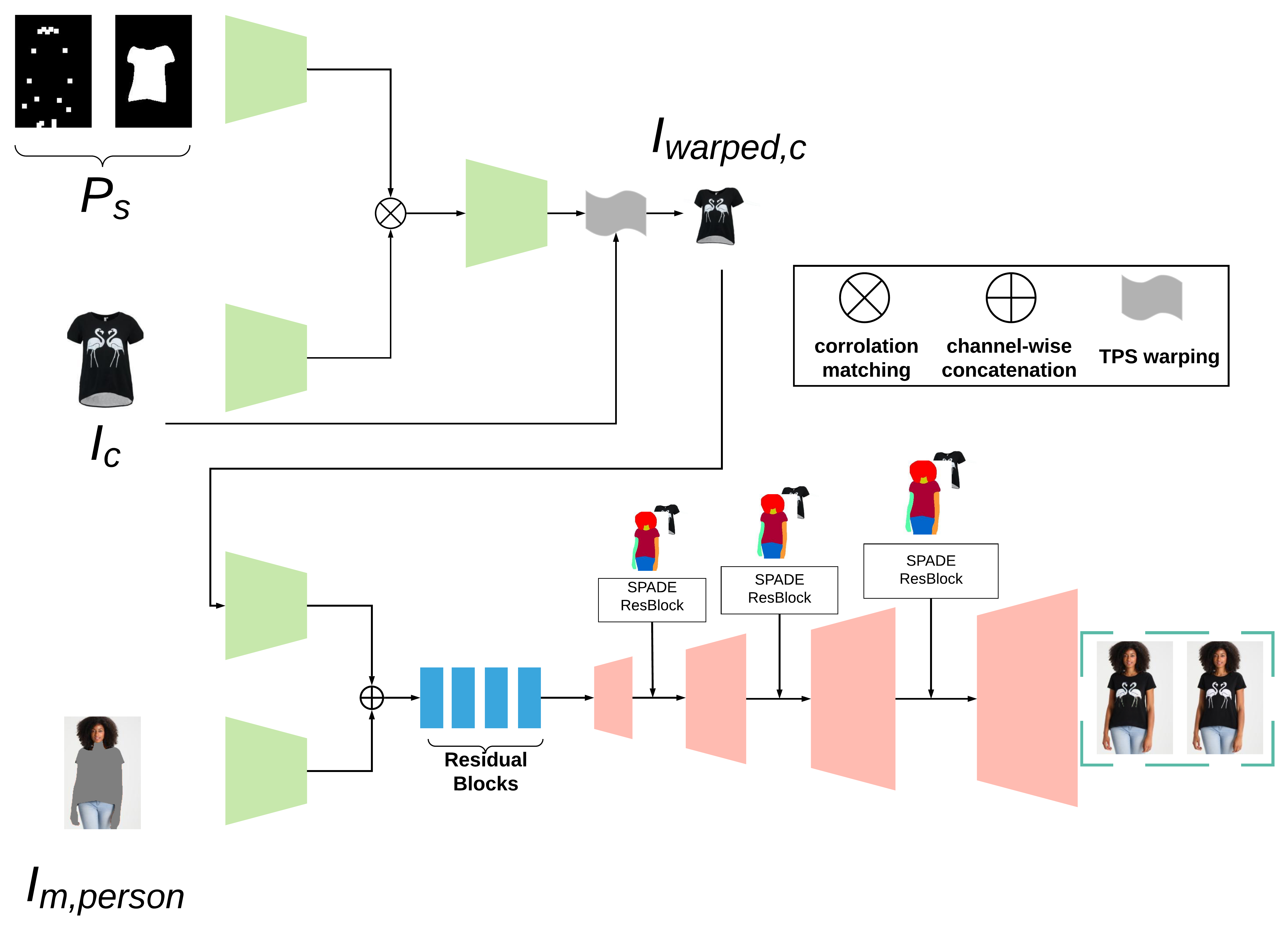}
\caption{Appearance transfer network architecture }
    \label{app_1}
\end{figure*}
Inspired by the current best architectures adopted for image-inpainting tasks  \cite{isola2017image,park2019semantic,ronneberger2015u}, we use an encoder-decoder structure for the generator network. The encoder is composed of standard convolutional layers with $3\times 3$ kernel sizes. The feature map is down-sampled with $2$ stride kernels for $5$ times, and instance normalization layers \cite{ulyanov2016instance} are utilized for normalizing the feature maps.  We utilize Leaky ReLU as a non-linear function applied to feature maps. The encoder is followed by $4$ residual blocks \cite{he2016deep} that serve as bottleneck and proceeds with a decoder that consists of convolutional residual blocks and nearest interpolation for up-sampling to the desired spatial resolution. It should be noted that the predicted semantic of the pixels falling out of the masked region (the gray region in Fig. \ref{seg_2}) and pixels associated with hands, legs, shoes and bottom clothing inside the masked region is replaced from that of the input segmentation map. Consequently, the identity of the reference person and the body parts that should not be subjected to any changes during the transference process are preserved.  Furthermore, the attention of the generator network is guided toward the areas needing manipulation during training. The generator is trained adversarially with a discriminator having PatchGan \cite{isola2017image} structure. Similar to the generator, the kernel sizes are $3\times 3$ and instance normalization and leaky ReLU are applied after convoluted feature maps in the discriminator network. 
The loss function for training the generator network is computed as:
\begin{equation}\label{eq:1}
L_{G} = \gamma_1 L_{parsing}+ \gamma_2 L_{per-pixel} - E[D(\hat{I}_{seg})]
\end{equation}
and that of discriminator is:
\begin{equation}\label{eq:2}
\begin{aligned}
L_{D} = E[max(0,1-D(I_{seg}))]+\\
E[max(0,1+D(\hat{I}_{seg}))]
+ \gamma_3 L_{GP} 
\end{aligned}
\end{equation}
In eq. \ref{eq:1}, $L_{parsing}$ corresponds to parsing loss \cite{gong2017look} and $L_{per-pixel}$ is the $L_1$ distance between the ground truth segmentation map $I_{seg}$ and generated map $\hat{I}_{seg}$ \cite{dong2019towards} defined as follow:
\begin{equation}\label{eq:3}
L_{per-pixel} = \frac{\lVert I_{seg} - \hat{I}_{seg} \rVert_1}{N}
\end{equation}
where $N$ denotes the number of pixels falling inside the masked region. Furthermore, $L_{GP}$ represents the gradient penalty loss \cite{gulrajani2017improved} that impacts the quality of the generated map significantly and makes the training process stable defined as:
\begin{equation}\label{eq:4}
L_{GP} = E[(\lVert \nabla_x D(x) \rVert_2 - 1)^{2}]
\end{equation}
where $x$ is a data point uniformly sampled from the line connecting $I_{seg}$ and $\hat{I}_{seg}$. In our work we set the values of hyper-parameters to be $\gamma_1 = 15$, $\gamma_2 = 20$ and $\gamma_3 = 10$.

\subsection{Appearance transfer Network}
\begin{figure*}[ht]
    \centering
       \includegraphics[height=3.7in]{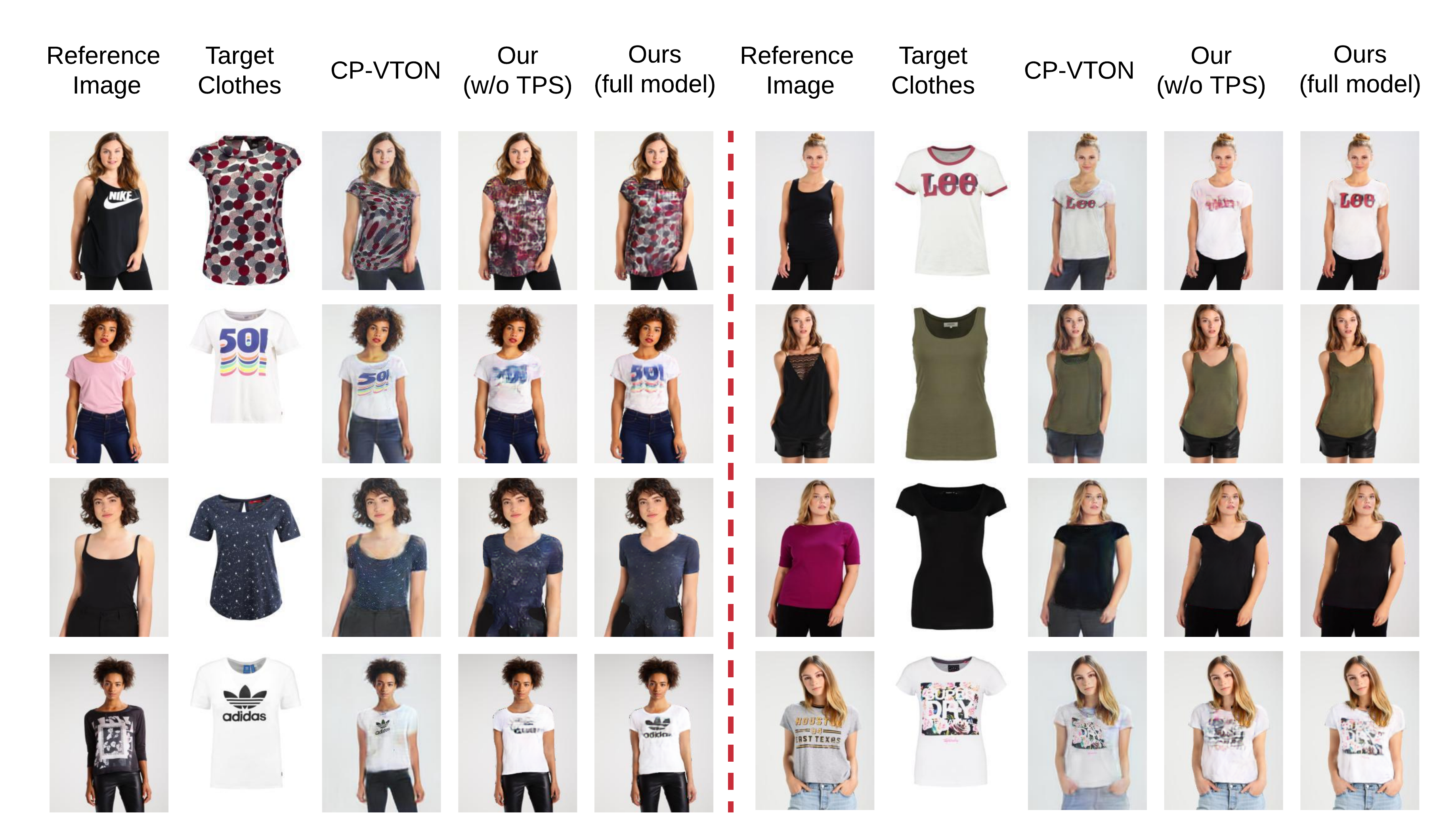}
\caption{This figure compares the quality of synthesized images by CP-VTON and GarmentGAN in terms of the sharpness and the fidelity in transferring small scale textural information from the target fashion item onto the body.}
    \label{robustness}
\end{figure*}
The appearance transfer network takes the generated segmentation maps, target clothing, and body shape information as inputs and generates an image portraying the reference person wearing the target clothing ($\hat{I}_{person}$). The overview of the proposed architecture is illustrated in Fig. \ref{app_1} . Similar to the shape transfer network, the proposed architecture consists of an adversarially  trained encoder-decoder generator. The input RGB-colorspace image ($I_{m, person}\in R^{H\times W\times 3}$) illustrated in Fig. \ref{app_1} is created by masking the regions corresponding to arms, torso and top clothes on the reference image ($I_{person}$). Inspired by \cite{wang2018toward}, a geometric alignment module is used (shown in Fig.\ref{app_1}.) to warp the desired clothing item ($I_c$) using a thin-plate spline (tps) transformer such that it matches the pose of reference person geometrically. This module extracts high-level information of person representation ($P_s$) and target clothing item ($I_c$) and uses the resulting feature maps to estimate the parameters of tps transformers ($\theta$). The estimated parameters are used to generate a warped clothing item ($I_{warped,c}$) that roughly aligns with the person's body according to eq. \ref{eq:5} .
\begin{equation}\label{eq:5}
I_{warped,c} = TPS_{\theta}(I_c)
\end{equation}

Furthermore, down-sampling and up-sampling the feature maps are conducted through 2-stride convolution and nearest interpolation, respectively. We use instance and spectral normalization \cite{miyato2018spectral} that together stabilize the training process. Additionally, we employ a SPADE-style normalization layer proposed by \cite{park2019semantic} to more accurately transfer spatial information. During the training phase, the semantic segmentation maps of the reference person $I_{seg}$ and warped cloth ($I_{warped,c}$) are concatenated in a channel-wise manner and fed into the SPADE layer. The  mean and variance of these feature maps are estimated at each layer in the decoder. 
\begin{equation}\label{eq:6}
\hat{I}_{person} = G_{appearance}(I_{m,person},I_c,P_s,I_{seg})
\end{equation}
We use multi-scale SN-PatchGan \cite{wang2018high} for the discriminator network that leads to generation of high quality RGB images. Hence, the loss function for the proposed network is:

\begin{equation}\label{eq:7}
\begin{aligned}
L_G = \alpha_1 L_{TPS} + \alpha_2 L_{per-pixel} + \alpha_3 L_{percept} \\
+ \alpha_4 L_{feat} - E[D(\hat{I}_{person})]  
\end{aligned}
\end{equation}
In eq. \ref{eq:7}, $L_{TPS}$ is the loss associated with Geometric Alignment module defined as:
\begin{equation}\label{eq:8}
L_{TPS} = E[\lVert I_{warped,c} - I_{worn,c} \rVert_1]
\end{equation}
where $I_{worn,c}$ denotes fashion item worn by reference person. In this equation, $L_{per-pixel}$ is $L_1$ distance between $\hat{I}_{person}$ and $I_{person}$ and $L_{percept}$ and $L_{feat}$ denote perceptual \cite{johnson2016perceptual} and feature matching \cite{wang2018high} losses, respectively. The loss function for the discriminator is:
\begin{equation}\label{eq:9}
\begin{aligned}
L_D = E[max(0,1-D(I_{person}))]+ \\
E[max(0,1+D(\hat{I}_{person}))] + \beta L_{GP}
\end{aligned}
\end{equation}

During the training, the values $\alpha$ and $\beta$ are set to $10$.
\begin{figure*}[ht]
    \centering
       \includegraphics[height=3.3in]{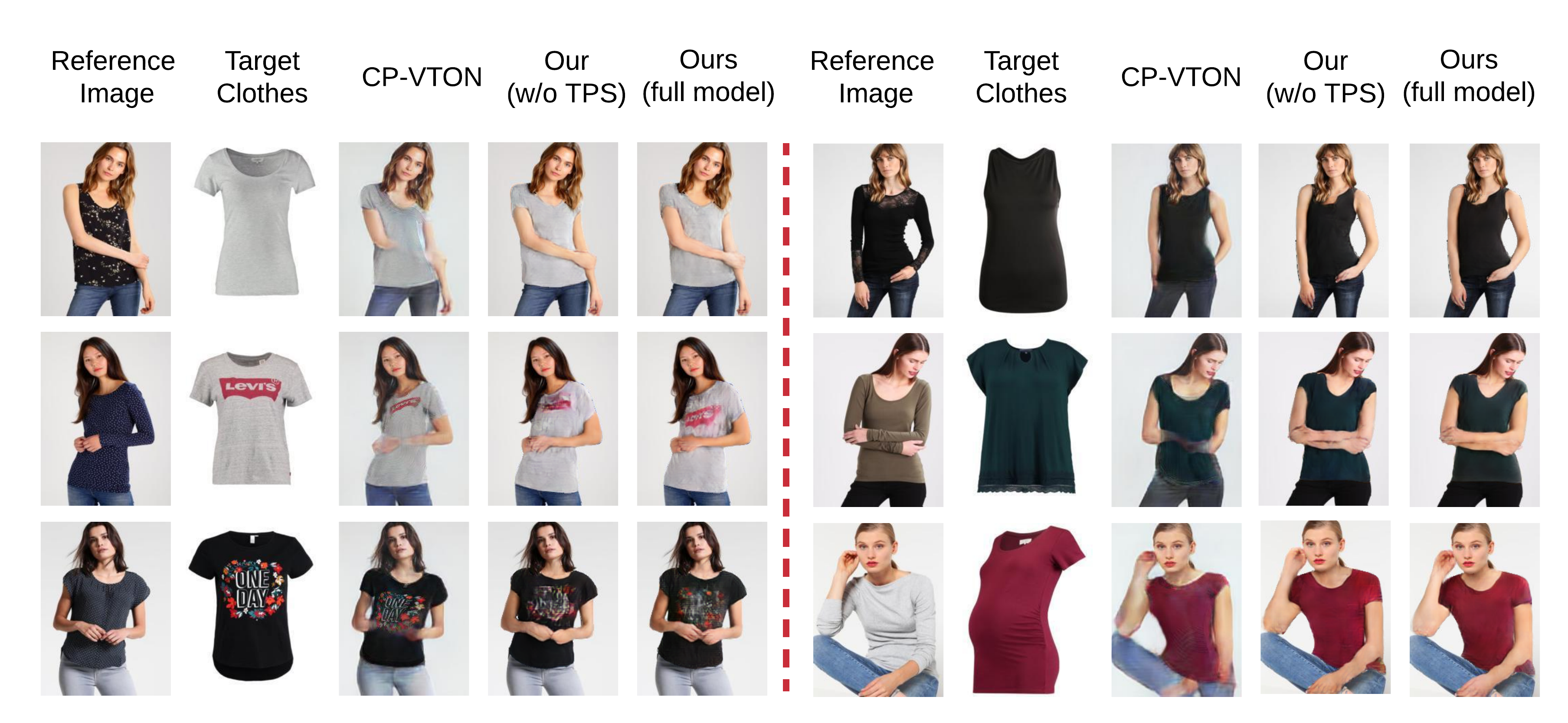}
\caption{This figure shows GarmentGAN's robust performance in the cases where the reference body has a complex pose. The texture, color and style of the clothing item are transferred successfully for the cases in which arms occlude the upper body.  }
    \label{occlusion}
\end{figure*}

\section{Experiments and discussion}

In this section, we describe the process and data used to evaluate the performance of the proposed algorithm. We also compare the results of our method with the state-of-the-art method (CP-VTON \cite{wang2018toward}) qualitatively and quantitatively. We also provide a comprehensive explanation as to why GarmentGAN outperforms CP-VTON so remarkably in various aspects.
\begin{figure*}[h]
    \centering
       \includegraphics[height=4.3in]{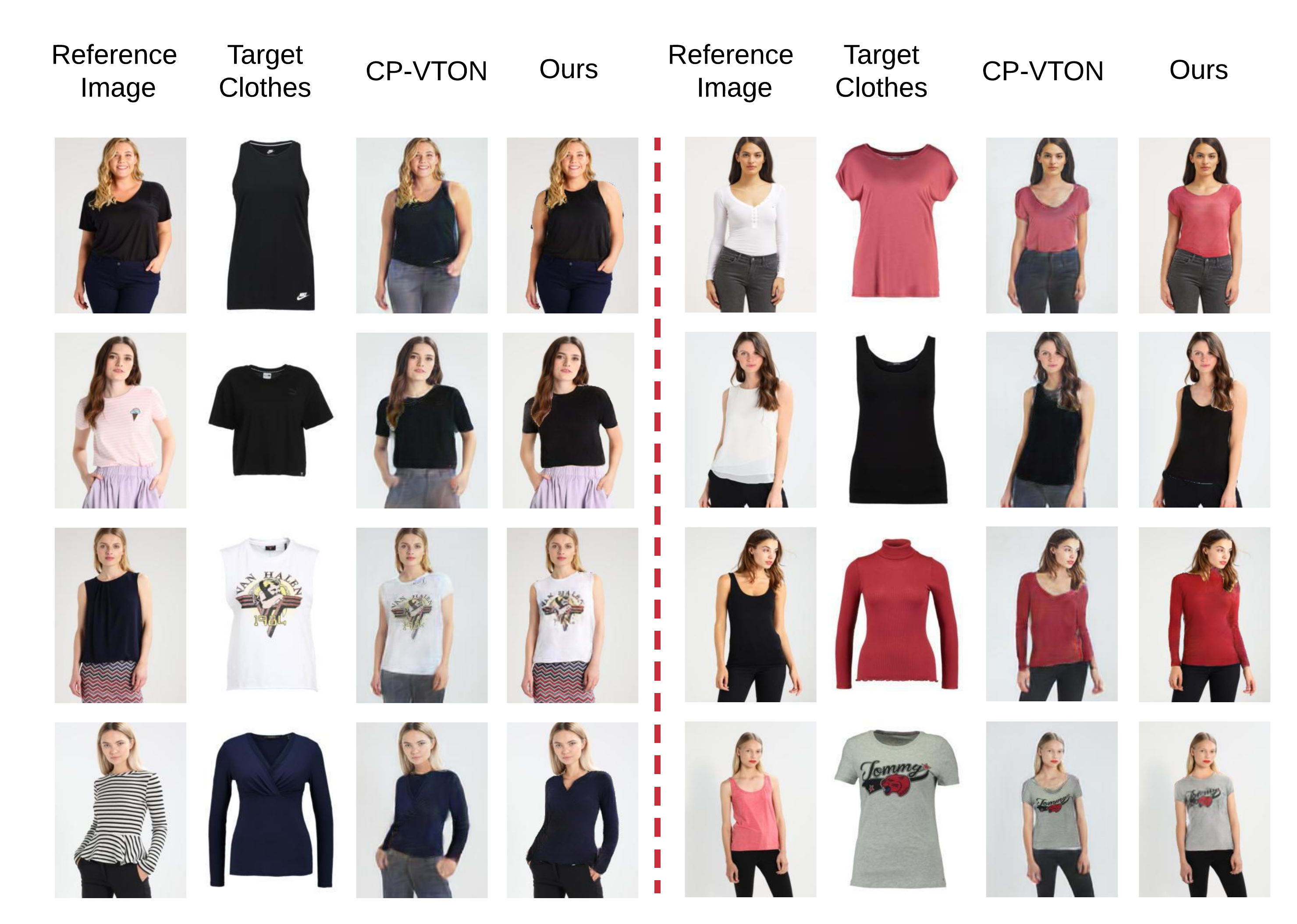}
\caption{In our method, the body parts and clothing items that should not be subjected to any changes during the garment transfer task are preserved successfully.  }
    \label{ID}
\end{figure*}

\subsection{Evaluation Dataset}
To show the performance of the proposed method, we used the dataset collected by \cite{han2018viton} to conduct all experiments. This dataset contains front-view images of women paired with the corresponding image of top clothing. Following \cite{wang2018toward}, the dataset is split into $14221$ training and $2032$ validation set, respectively.

\subsection{Evaluation Metrics}
We use two popular measures to evaluate the performance of our model and compare the quality and realism of the generated image with state-of-the-art methods and various implementations of our work; (i) Inception Score (\textbf{IS}) \cite{salimans2016improved} and (ii) Fr\'echet Inception Distance (\textbf{FID}) \cite{heusel2017gans}. \textbf{IS} uses Inception-v3 network \cite{szegedy2016rethinking} pre-trained on ImageNet \cite{deng2009imagenet} dataset to compare the conditional label distribution and marginal label distribution of generated images. Higher \textbf{IS} indicates that produced images are diverse, and each one has a more meaningful object in it. On the other hand, \textbf{FID} computes the Fr\'echet distance between the distribution of the real and synthesized dataset. For this purpose, a multivariate Gaussian distribution is assigned to the feature maps of synthesized and real datasets passed through a pre-trained Inception-v3 network, and the distance between these two distributions is computed. Hence, a smaller distance induces more similarity between real and generated datasets.

\subsection{Visual comparison}
We perform a visual comparison between CP-VTON \cite{wang2018toward}, our proposed model without Geometric Alignment module (ours without TPS), and our full model. Furthermore, we discuss the quality of generated images from various aspects below:

\subsubsection{Sharpness of the synthesized image and fidelity to appearance transfer}

Figure \ref{robustness} illustrates a comparison between the images generated by our model and CP-VTON. The images produced by our model have promising quality (sharp images) and the details in the appearance of the clothing item (e.g. texture, color, logo) have been properly transferred onto the reference body. However, CP-VTON synthesizes blurry images where the boundaries are not sharp and contain artifacts. The gained superiority of GarmentGAN is due to the disentanglement of shape and appearance transfer tasks. As shown in this work, the boundaries of new clothing items and various body parts are identified in the first stage and serve as a guide for transferring the target garment accurately to the most reasonable location in the latter stage. This disentanglement helps to alleviate the boundary artifact problem and avoid blurring during synthesis. It should also be noted that the appearance transfer is significantly enhanced due to the geometric alignment module used in the network compared to the state-of-the-art method. In our method, the warping module is trained in an End-to-End fashion with the appearance transfer network, while this module is trained separately in CP-VITON which leads to poor transfer performance. Finally, we tried the shape transference stage without the warping module and the results indicate that the details of the target clothing item cannot be reconstructed convincingly in the generated image.

\subsubsection{Complex Pose Performance}

We also evaluate the performance of the model in cases where the pose of the reference person is complicated, such as when arms and body occlude each other in Fig. \ref{occlusion}. The generated images show that our method is more robust to arbitrary pose (e.g. when arms occlude the upper body or vice versa) compared to CP-VTON. This improvement is achieved in part due to the way the input segmentation map for the shape transfer network is masked. The pixels associated with the hands are identified and remain unchanged as they are passed through GarmentGAN's model. This practice helps the network to generate segmentation maps in which hands have a reasonable layout with respect to the body and preserves hand gestures through the transfer process.

\subsubsection{Preserving Identity in the Reference Image}

According to Fig. \ref{ID}, GarmentGAN maintains the identity of the person in the reference image and the details of the bottom clothing are preserved in produced images. However, the bottom clothes worn by the person are not reconstructed in the synthesized image by CP-VTON. Moreover, many crucial structural details of the target clothing, such as collar shapes, are not transferred properly using CP-VTON. Our results clearly show that GarmentGAN is able to reconstruct the geometrical features of the target clothing item faithfully

\subsection{Quantitative comparison}
To compare performance of GarmentGAN with that of the current state-of-the-art technique, the \textbf{IS} and \textbf{FID} scores are computed and presented in Table. \ref{comparison}. The results show that our proposed method outperforms CP-VTON in terms of realism and quality of the generated images.

\begin{table}
\centering
\begin{tabular}{lcc}
\hline
Model & \textbf{IS} & \textbf{FID} \\
\hline \hline
CP-VTON  & $2.636 \pm 0.077$ & $23.085$ \\
Our model w/o TPS & $2.723 \pm 0.083$ & $17.408$ \\
Our model (Full) & $\textbf{2.774} \pm \textbf{0.082}$ & $\textbf{16.578}$\\
\hline
\end{tabular}
\captionof{table}{Performance comparison between CP-VTON, GarmentGAN without TPS, and full GarmentGAN}
\label{comparison}
\end{table}

\section{Conclusion}
We introduce GarmentGAN, a generative adversarial treatment of the challenging garment transfer task in unconstrained images. This end-to-end trained model is shown to synthesize high quality images and robustly transfer photographic characteristics of clothing. Furthermore, this approach alleviates problems related to complexity of body pose, garment shape, and occlusions - capabilities which are achieved by employing novel network design and adversarial training schemes. Extensive model performance evaluations have been conducted and  results show significant improvements over current state-of-the-art image-to-image fashion transfer pipelines.

\bibliographystyle{acmsiggraph}
\nocite{*}
\bibliography{template}

\begin{thebibliography}{\protect\citename{Ronneberger et~al\mbox{.} }2015}

\bibitem[\protect\citename{Chen et~al\mbox{.} }2018]{chen2018cascaded}
{\sc Chen, Y., Wang, Z., Peng, Y., Zhang, Z., Yu, G., and Sun, J.}
\newblock 2018.
\newblock Cascaded pyramid network for multi-person pose estimation.
\newblock In {\em Proceedings of the IEEE Conference on Computer Vision and
  Pattern Recognition}, 7103--7112.

\bibitem[\protect\citename{Chen et~al\mbox{.} }2019]{chen2019unpaired}
{\sc Chen, X., Song, J., and Hilliges, O.}
\newblock 2019.
\newblock Unpaired pose guided human image generation.
\newblock {\em arXiv preprint arXiv:1901.02284\/}.

\bibitem[\protect\citename{Deng et~al\mbox{.} }2009]{deng2009imagenet}
{\sc Deng, J., Dong, W., Socher, R., Li, L.-J., Li, K., and Fei-Fei, L.}
\newblock 2009.
\newblock Imagenet: A large-scale hierarchical image database.
\newblock In {\em 2009 IEEE conference on computer vision and pattern
  recognition}, Ieee, 248--255.

\bibitem[\protect\citename{Dong et~al\mbox{.} }2019]{dong2019towards}
{\sc Dong, H., Liang, X., Wang, B., Lai, H., Zhu, J., and Yin, J.}
\newblock 2019.
\newblock Towards multi-pose guided virtual try-on network.
\newblock {\em arXiv preprint arXiv:1902.11026\/}.

\bibitem[\protect\citename{Gong et~al\mbox{.} }2017]{gong2017look}
{\sc Gong, K., Liang, X., Zhang, D., Shen, X., and Lin, L.}
\newblock 2017.
\newblock Look into person: Self-supervised structure-sensitive learning and a
  new benchmark for human parsing.
\newblock In {\em Proceedings of the IEEE Conference on Computer Vision and
  Pattern Recognition}, 932--940.

\bibitem[\protect\citename{Gong et~al\mbox{.} }2018]{gong2018instance}
{\sc Gong, K., Liang, X., Li, Y., Chen, Y., Yang, M., and Lin, L.}
\newblock 2018.
\newblock Instance-level human parsing via part grouping network.
\newblock In {\em Proceedings of the European Conference on Computer Vision
  (ECCV)}, 770--785.

\bibitem[\protect\citename{Goodfellow et~al\mbox{.}
  }2014]{goodfellow2014generative}
{\sc Goodfellow, I., Pouget-Abadie, J., Mirza, M., Xu, B., Warde-Farley, D.,
  Ozair, S., Courville, A., and Bengio, Y.}
\newblock 2014.
\newblock Generative adversarial nets.
\newblock In {\em Advances in neural information processing systems},
  2672--2680.

\bibitem[\protect\citename{Gulrajani et~al\mbox{.}
  }2017]{gulrajani2017improved}
{\sc Gulrajani, I., Ahmed, F., Arjovsky, M., Dumoulin, V., and Courville,
  A.~C.}
\newblock 2017.
\newblock Improved training of wasserstein gans.
\newblock In {\em Advances in neural information processing systems},
  5767--5777.

\bibitem[\protect\citename{Han et~al\mbox{.} }2018]{han2018viton}
{\sc Han, X., Wu, Z., Wu, Z., Yu, R., and Davis, L.~S.}
\newblock 2018.
\newblock Viton: An image-based virtual try-on network.
\newblock In {\em Proceedings of the IEEE Conference on Computer Vision and
  Pattern Recognition}, 7543--7552.

\bibitem[\protect\citename{Han et~al\mbox{.} }2019]{han2019compatible}
{\sc Han, X., Wu, Z., Huang, W., Scott, M.~R., and Davis, L.~S.}
\newblock 2019.
\newblock Compatible and diverse fashion image inpainting.
\newblock {\em arXiv preprint arXiv:1902.01096\/}.

\bibitem[\protect\citename{He et~al\mbox{.} }2016]{he2016deep}
{\sc He, K., Zhang, X., Ren, S., and Sun, J.}
\newblock 2016.
\newblock Deep residual learning for image recognition.
\newblock In {\em Proceedings of the IEEE conference on computer vision and
  pattern recognition}, 770--778.

\bibitem[\protect\citename{Heusel et~al\mbox{.} }2017]{heusel2017gans}
{\sc Heusel, M., Ramsauer, H., Unterthiner, T., Nessler, B., and Hochreiter,
  S.}
\newblock 2017.
\newblock Gans trained by a two time-scale update rule converge to a local nash
  equilibrium.
\newblock In {\em Advances in Neural Information Processing Systems},
  6626--6637.

\bibitem[\protect\citename{Isola et~al\mbox{.} }2017]{isola2017image}
{\sc Isola, P., Zhu, J.-Y., Zhou, T., and Efros, A.~A.}
\newblock 2017.
\newblock Image-to-image translation with conditional adversarial networks.
\newblock In {\em Proceedings of the IEEE conference on computer vision and
  pattern recognition}, 1125--1134.

\bibitem[\protect\citename{Issenhuth et~al\mbox{.} }2019]{issenhuth2019end}
{\sc Issenhuth, T., Mary, J., and Calauzennes, C.}
\newblock 2019.
\newblock End-to-end learning of geometric deformations of feature maps for
  virtual try-on.
\newblock {\em arXiv preprint arXiv:1906.01347\/}.

\bibitem[\protect\citename{Jetchev and Bergmann }2017]{jetchev2017conditional}
{\sc Jetchev, N., and Bergmann, U.}
\newblock 2017.
\newblock The conditional analogy gan: Swapping fashion articles on people
  images.
\newblock In {\em Proceedings of the IEEE International Conference on Computer
  Vision}, 2287--2292.

\bibitem[\protect\citename{Johnson et~al\mbox{.} }2016]{johnson2016perceptual}
{\sc Johnson, J., Alahi, A., and Fei-Fei, L.}
\newblock 2016.
\newblock Perceptual losses for real-time style transfer and super-resolution.
\newblock In {\em European conference on computer vision}, Springer, 694--711.

\bibitem[\protect\citename{Kim et~al\mbox{.} }2017]{kim2017learning}
{\sc Kim, T., Cha, M., Kim, H., Lee, J.~K., and Kim, J.}
\newblock 2017.
\newblock Learning to discover cross-domain relations with generative
  adversarial networks.
\newblock In {\em Proceedings of the 34th International Conference on Machine
  Learning-Volume 70}, JMLR. org, 1857--1865.

\bibitem[\protect\citename{Kingma and Ba }2014]{kingma2014adam}
{\sc Kingma, D.~P., and Ba, J.}
\newblock 2014.
\newblock Adam: A method for stochastic optimization.
\newblock {\em arXiv preprint arXiv:1412.6980\/}.

\bibitem[\protect\citename{Kubo et~al\mbox{.} }2018]{kubo2018generative}
{\sc Kubo, S., Iwasawa, Y., and Matsuo, Y.}, 2018.
\newblock Generative adversarial network-based virtual try-on with clothing
  region.

\bibitem[\protect\citename{Lahner et~al\mbox{.} }2018]{lahner2018deepwrinkles}
{\sc Lahner, Z., Cremers, D., and Tung, T.}
\newblock 2018.
\newblock Deepwrinkles: Accurate and realistic clothing modeling.
\newblock In {\em Proceedings of the European Conference on Computer Vision
  (ECCV)}, 667--684.

\bibitem[\protect\citename{Lassner et~al\mbox{.} }2017]{lassner2017generative}
{\sc Lassner, C., Pons-Moll, G., and Gehler, P.~V.}
\newblock 2017.
\newblock A generative model of people in clothing.
\newblock In {\em Proceedings of the IEEE International Conference on Computer
  Vision}, 853--862.

\bibitem[\protect\citename{Lin et~al\mbox{.} }2014]{lin2014microsoft}
{\sc Lin, T.-Y., Maire, M., Belongie, S., Hays, J., Perona, P., Ramanan, D.,
  Doll{\'a}r, P., and Zitnick, C.~L.}
\newblock 2014.
\newblock Microsoft coco: Common objects in context.
\newblock In {\em European conference on computer vision}, Springer, 740--755.

\bibitem[\protect\citename{Liu et~al\mbox{.} }2019]{liu2019swapgan}
{\sc Liu, Y., Chen, W., Liu, L., and Lew, M.~S.}
\newblock 2019.
\newblock Swapgan: A multistage generative approach for person-to-person
  fashion style transfer.
\newblock {\em IEEE Transactions on Multimedia\/}.

\bibitem[\protect\citename{Ma et~al\mbox{.} }2017]{ma2017pose}
{\sc Ma, L., Jia, X., Sun, Q., Schiele, B., Tuytelaars, T., and Van~Gool, L.}
\newblock 2017.
\newblock Pose guided person image generation.
\newblock In {\em Advances in Neural Information Processing Systems}, 406--416.

\bibitem[\protect\citename{Ma et~al\mbox{.} }2018]{ma2018disentangled}
{\sc Ma, L., Sun, Q., Georgoulis, S., Van~Gool, L., Schiele, B., and Fritz, M.}
\newblock 2018.
\newblock Disentangled person image generation.
\newblock In {\em Proceedings of the IEEE Conference on Computer Vision and
  Pattern Recognition}, 99--108.

\bibitem[\protect\citename{Miyato et~al\mbox{.} }2018]{miyato2018spectral}
{\sc Miyato, T., Kataoka, T., Koyama, M., and Yoshida, Y.}
\newblock 2018.
\newblock Spectral normalization for generative adversarial networks.
\newblock {\em arXiv preprint arXiv:1802.05957\/}.

\bibitem[\protect\citename{Park et~al\mbox{.} }2019]{park2019semantic}
{\sc Park, T., Liu, M.-Y., Wang, T.-C., and Zhu, J.-Y.}
\newblock 2019.
\newblock Semantic image synthesis with spatially-adaptive normalization.
\newblock In {\em Proceedings of the IEEE Conference on Computer Vision and
  Pattern Recognition}, 2337--2346.

\bibitem[\protect\citename{Pons-Moll et~al\mbox{.} }2017]{pons2017clothcap}
{\sc Pons-Moll, G., Pujades, S., Hu, S., and Black, M.~J.}
\newblock 2017.
\newblock Clothcap: Seamless 4d clothing capture and retargeting.
\newblock {\em ACM Transactions on Graphics (TOG) 36}, 4, 73.

\bibitem[\protect\citename{Pumarola et~al\mbox{.}
  }2018]{pumarola2018unsupervised}
{\sc Pumarola, A., Agudo, A., Sanfeliu, A., and Moreno-Noguer, F.}
\newblock 2018.
\newblock Unsupervised person image synthesis in arbitrary poses.
\newblock In {\em Proceedings of the IEEE Conference on Computer Vision and
  Pattern Recognition}, 8620--8628.

\bibitem[\protect\citename{Raj et~al\mbox{.} }2018]{raj2018swapnet}
{\sc Raj, A., Sangkloy, P., Chang, H., Lu, J., Ceylan, D., and Hays, J.}
\newblock 2018.
\newblock Swapnet: Garment transfer in single view images.
\newblock In {\em Proceedings of the European Conference on Computer Vision
  (ECCV)}, 666--682.

\bibitem[\protect\citename{Reed et~al\mbox{.} }2016]{reed2016generative}
{\sc Reed, S., Akata, Z., Yan, X., Logeswaran, L., Schiele, B., and Lee, H.}
\newblock 2016.
\newblock Generative adversarial text to image synthesis.
\newblock {\em arXiv preprint arXiv:1605.05396\/}.

\bibitem[\protect\citename{Ronneberger et~al\mbox{.} }2015]{ronneberger2015u}
{\sc Ronneberger, O., Fischer, P., and Brox, T.}
\newblock 2015.
\newblock U-net: Convolutional networks for biomedical image segmentation.
\newblock In {\em International Conference on Medical image computing and
  computer-assisted intervention}, Springer, 234--241.

\bibitem[\protect\citename{Salimans et~al\mbox{.} }2016]{salimans2016improved}
{\sc Salimans, T., Goodfellow, I., Zaremba, W., Cheung, V., Radford, A., and
  Chen, X.}
\newblock 2016.
\newblock Improved techniques for training gans.
\newblock In {\em Advances in neural information processing systems},
  2234--2242.

\bibitem[\protect\citename{Siarohin et~al\mbox{.}
  }2018]{siarohin2018deformable}
{\sc Siarohin, A., Sangineto, E., Lathuili{\`e}re, S., and Sebe, N.}
\newblock 2018.
\newblock Deformable gans for pose-based human image generation.
\newblock In {\em Proceedings of the IEEE Conference on Computer Vision and
  Pattern Recognition}, 3408--3416.

\bibitem[\protect\citename{Szegedy et~al\mbox{.} }2016]{szegedy2016rethinking}
{\sc Szegedy, C., Vanhoucke, V., Ioffe, S., Shlens, J., and Wojna, Z.}
\newblock 2016.
\newblock Rethinking the inception architecture for computer vision.
\newblock In {\em Proceedings of the IEEE conference on computer vision and
  pattern recognition}, 2818--2826.

\bibitem[\protect\citename{Ulyanov et~al\mbox{.} }2016]{ulyanov2016instance}
{\sc Ulyanov, D., Vedaldi, A., and Lempitsky, V.}
\newblock 2016.
\newblock Instance normalization: The missing ingredient for fast stylization.
\newblock {\em arXiv preprint arXiv:1607.08022\/}.

\bibitem[\protect\citename{Wang et~al\mbox{.} }2018a]{wang2018toward}
{\sc Wang, B., Zheng, H., Liang, X., Chen, Y., Lin, L., and Yang, M.}
\newblock 2018.
\newblock Toward characteristic-preserving image-based virtual try-on network.
\newblock In {\em Proceedings of the European Conference on Computer Vision
  (ECCV)}, 589--604.

\bibitem[\protect\citename{Wang et~al\mbox{.} }2018b]{wang2018high}
{\sc Wang, T.-C., Liu, M.-Y., Zhu, J.-Y., Tao, A., Kautz, J., and Catanzaro,
  B.}
\newblock 2018.
\newblock High-resolution image synthesis and semantic manipulation with
  conditional gans.
\newblock In {\em Proceedings of the IEEE conference on computer vision and
  pattern recognition}, 8798--8807.

\bibitem[\protect\citename{Yan et~al\mbox{.} }2017]{yan2017skeleton}
{\sc Yan, Y., Xu, J., Ni, B., Zhang, W., and Yang, X.}
\newblock 2017.
\newblock Skeleton-aided articulated motion generation.
\newblock In {\em Proceedings of the 25th ACM international conference on
  Multimedia}, ACM, 199--207.

\bibitem[\protect\citename{Yi et~al\mbox{.} }2017]{yi2017dualgan}
{\sc Yi, Z., Zhang, H., Tan, P., and Gong, M.}
\newblock 2017.
\newblock Dualgan: Unsupervised dual learning for image-to-image translation.
\newblock In {\em Proceedings of the IEEE international conference on computer
  vision}, 2849--2857.

\bibitem[\protect\citename{Zhang et~al\mbox{.} }2017]{zhang2017detailed}
{\sc Zhang, C., Pujades, S., Black, M.~J., and Pons-Moll, G.}
\newblock 2017.
\newblock Detailed, accurate, human shape estimation from clothed 3d scan
  sequences.
\newblock In {\em Proceedings of the IEEE Conference on Computer Vision and
  Pattern Recognition}, 4191--4200.

\bibitem[\protect\citename{Zhu et~al\mbox{.} }2017a]{zhu2017unpaired}
{\sc Zhu, J.-Y., Park, T., Isola, P., and Efros, A.~A.}
\newblock 2017.
\newblock Unpaired image-to-image translation using cycle-consistent
  adversarial networks.
\newblock In {\em Proceedings of the IEEE international conference on computer
  vision}, 2223--2232.

\bibitem[\protect\citename{Zhu et~al\mbox{.} }2017b]{zhu2017your}
{\sc Zhu, S., Urtasun, R., Fidler, S., Lin, D., and Change~Loy, C.}
\newblock 2017.
\newblock Be your own prada: Fashion synthesis with structural coherence.
\newblock In {\em Proceedings of the IEEE International Conference on Computer
  Vision}, 1680--1688.

\end{thebibliography}
\end{document}